\documentclass[10pt,twocolumn,letterpaper]{article}

\usepackage{cvpr}
\usepackage{times}
\usepackage{epsfig}
\usepackage{graphicx}
\usepackage{amsmath}
\usepackage{amssymb}

\usepackage{pifont}% http://ctan.org/pkg/pifont
\newcommand{\cmark}{\ding{51}}%
\newcommand{\xmark}{\ding{55}}%
\usepackage{color}
\usepackage[breaklinks=true,bookmarks=false]{hyperref}

\cvprfinalcopy % *** Uncomment this line for the final submission

\def\cvprPaperID{4186} % *** Enter the CVPR Paper ID here

\ifcvprfinal\fi
\begin{document}

%%%%%%%%% TITLE
\title{Unsupervised Learning of Monocular Depth Estimation and Visual Odometry with Deep Feature Reconstruction
}

\author{Huangying Zhan{$^{1,2}$}, Ravi Garg{$^{1,2}$}, Chamara Saroj Weerasekera{$^{1,2}$}, Kejie Li{$^{1,2}$}, Harsh Agarwal{$^3$}, Ian Reid{$^{1,2}$}\\
	$^{1}$The University of Adelaide\\
	$^{2}$Australian Centre for Robotic Vision\\
	$^{3}$Indian Institute of Technology (BHU)\\
	\tt\small\{huangying.zhan, ravi.garg, chamara.weerasekera, kejie.li, ian.reid\}@adelaide.edu.au\\
	\tt\small harsh.agarwal.eee14@iitbhu.ac.in
}

\maketitle
\thispagestyle{empty}

%%%%%%%%% ABSTRACT
\begin{abstract}

Despite learning based methods showing promising results in single view depth estimation and visual odometry, most existing approaches treat the tasks in a supervised manner. 
Recent approaches to single view depth estimation explore the possibility of learning without full supervision via minimizing photometric error.
In this paper, we explore the use of stereo sequences for learning depth and visual odometry.
The use of stereo sequences enables the use of both spatial (between left-right pairs) and temporal (forward backward) photometric warp error, and constrains the scene depth and camera motion to be in a common, real-world scale. At test time our framework is able to estimate single view depth and two-view odometry from a monocular sequence. We also show how we can improve on a standard photometric warp loss by considering a warp of deep features. 
We show through extensive experiments that: (i) jointly training for single view depth and visual odometry improves depth prediction because of the additional constraint imposed on depths and achieves competitive results for visual odometry; (ii) deep feature-based warping loss improves upon simple photometric warp loss for both single view depth estimation and visual odometry. Our method outperforms existing learning based methods on the KITTI driving dataset in both tasks. The source code is available at \url{https://github.com/Huangying-Zhan/Depth-VO-Feat}.

\end{abstract}

\section{Introduction} \label{sec:intro}

\begin{figure}[h] 
\centering
    \includegraphics[width=1\columnwidth]{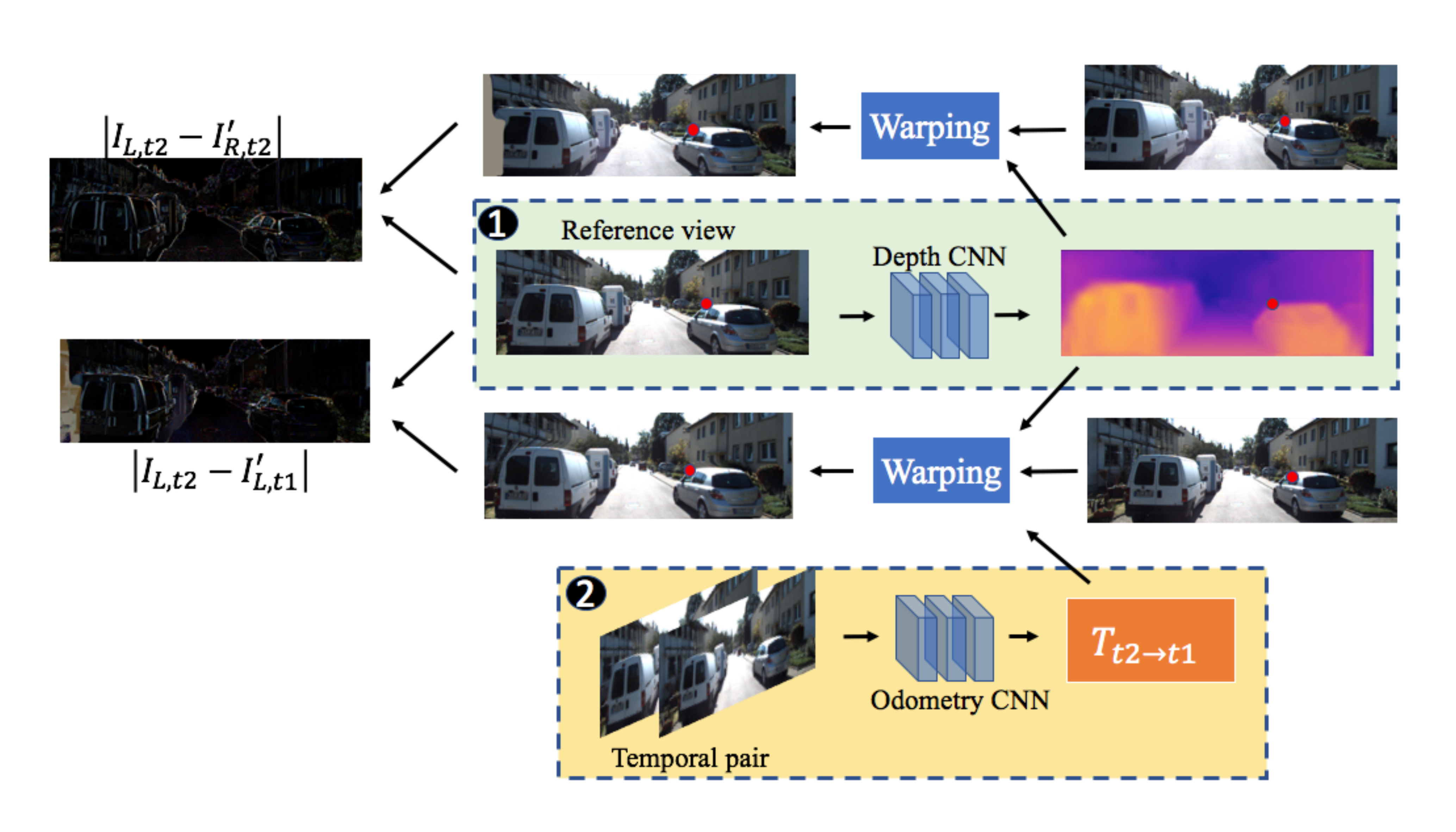}
    \caption{Training instance example. The known camera motion between stereo cameras $T_{L \rightarrow R}$ constrains the Depth CNN and Odometry CNN to predict depth and relative camera pose with actual scale.
    % \vspace{-10mm}
    }\label{fig:train_eg}
\end{figure}

Understanding the 3D structure of a scene from a single image is a fundamental question in machine perception. 
The related problem of inferring ego-motion from a sequence of images is likewise a fundamental problem in robotics, known as visual odometry estimation.
These two problems are crucial in robotic vision research since accurate estimation of depth and odometry based on images has many important applications, most notably for autonomous vehicles.

While both problems have been the subject of research in robotic vision since the origins of the discipline, with numerous geometric solutions proposed, in recent times a number of works have cast depth estimation and visual odometry as supervised learning problems \cite{agrawal2015seebymoving}\cite{eigen2014depth}\cite{liu2015depth}\cite{liu2016depth}. These methods attempt to predict depth or odometry using models that have been trained from a large dataset with ground truth data. 
However, these annotations are expensive to obtain, e.g. expensive laser or depth camera to collect depths. In a recent work Garg \textit{et al.}\cite{garg2016depth}  recognised that these tasks are amenable to an unsupervised framework where the authors propose to use photometric warp error as a self-supervised signal to train a convolutional neural network (ConvNet / CNN) for the single view depth estimation. Following \cite{garg2016depth}  methods like \cite{godard2016depth}\cite{kuznietsov2017semi}\cite{ye2017self} use the photometric error based supervision to learn depth estimators comparable to that of fully supervised methods.
Specifically, \cite{garg2016depth} and \cite{godard2016depth} use the photometric warp error between left-right images in a stereo pair to learn depth. 
Recognising the generality of the idea, \cite{zhou2017sfmlearner} uses monocular sequences to jointly train two neural networks for depth and odometry estimation. 
However, relying on the two frame visual odometry estimation framework, \cite{zhou2017sfmlearner} suffers from the per frame scale-ambiguity issue, in that an actual metric scaling of the camera translations is missing and only direction is known.
Having a good estimate of the translation scale per-frame is crucial for the success of any Simultaneous Localization and Mapping (SLAM) system. Accurate camera tracking in most monocular SLAM frameworks relies on keeping the scale consistency of the map across multiple images which is enforced using a single scale map. In absence of a global map for tracking, an expensive bundle adjustment over the per-frame scale parameter or additional assumptions like constant camera height from the already detected ground plane becomes essential for accurate tracking \cite{song2014robust}.

In this work, we propose a framework which jointly learns a single view depth estimator and monocular odometry estimator using stereo video sequences (as shown in Figure \ref{fig:train_eg})  for training. Our method can be understood as unsupervised learning for depth estimation and semi-supervised for pose which is known between stereo pairs. The use of stereo sequences enables the use of both spatial (between left-right pairs) and temporal (forward-backward) photometric warp error, and constrains the scene depth and camera motion to be in a common, real-world scale (set by the stereo baseline). Inference (i.e. depth and odometry estimation) without any scale ambiguity is then possible using a single camera for pure frame to frame VO estimation without any need for mapping.

Moreover, while the previous works have shown the efficacy of using the photometric warp error as a self-supervision signal, a simple warp of image intensities or colors carries its own assumptions about brightness/color consistency, and must also be accompanied by a regularization to generate ``sensible'' warps when the photometric information is ambiguous, such as in uniformly colored regions (see Sec.\ref{sec:method_featSuper}). We propose an additional deep feature reconstruction loss which takes contextual information into consideration rather than per pixel color matching alone. 

In summary, we make the following contributions: 
(i) an unsupervised framework for jointly learning a depth estimator and visual odometry estimator that does not suffer from the scale ambiguity; 
(ii) takes advantage of the full set of constraints available from spatial and temporal image pairs to improve upon prior art on single view depth estimations; 
(iii) produces the state-of-the-art frame-to-frame odometry results that significantly improve on \cite{zhou2017sfmlearner} and are on par with geometric methods; 
(iv) uses a novel feature reconstruction loss in addition to the color intensity based image reconstruction loss which improves the depth and odometry estimation accuracy significantly.

\section{Related Work} \label{sec:relWork}

Humans are capable of reasoning the relative depth of pixels in an image and perceive ego-motion given two images, but both single view depth estimation and two frame visual odometry are challenging problems. Avoiding visual learning, localization and 3D reconstruction in computer vision was considered a purely geometric problem for decades. 
While prior to deep learning graphical models based learning methods \cite{saxena2006depth}\cite{saxena2009make3d} were prevalent examples for single view reconstructions, methods based on the epipolar geometry were to the fore for two view odometry. While it is possible to estimate the relative pose between two frames based only on the data within those two frames up-to a scale (see e.g., \cite{LonguetHiggins81}, the ``gold-standard'' for geometric ego-motion estimation to date is based on a batch bundle adjustment of pose and scene structure \cite{triggs1999bundle}, or on online Visual SLAM technqiues \cite{davison2007monoslam}). After the surge of convolutional neural networks, both depth and visual odometry estimation problem have been attempted with deep learning methods.

\paragraph{Supervised methods}
Deep learning based depth estimation starts with Eigen \textit{et al.} \cite{eigen2014depth} which is the first work estimating depth with ConvNets. They used a multi-scale deep network and scale-invariant loss for depth estimation. 
Liu \textit{et al.}\cite{liu2015depth} \cite{liu2016depth} formulated depth estimation as a continuous conditional random field learning problem. 
Laina \textit{et al.} \cite{laina2016deeperdepth} proposed a residual network using fully convolutional architecture to model the mapping between monocular image and depth map. They also introduced reverse Huber loss and newly designed up-sampling modules. 
Kendell  \textit{et al.}\cite{kendall2017deepstereo} proposed an end-to-end learning framework to predict disparity from a stereo pair. In particular, they propose to use an explicit feature matching step as a layer in the network to create the cost-volume matching two images, which is then regularized to predict the state-of-the-art disparities for outdoor stereo sequences on KITTI dataset.

For odometry, Agrawal \textit{et al.} \cite{agrawal2015seebymoving} proposed a visual feature learning algorithm which aims at learning good visual features. Instead of learning features from a classification task (e.g. ImageNet\cite{ILSVRC15}), \cite{agrawal2015seebymoving} learns features from an ego-motion estimation task. The model is capable to estimate relative camera poses.
Wang \textit{et al.}\cite{wang2017deepvo} presented a recurrent ConvNet architecture for learning monocular odometry from video sequences.

Ummenhofer \textit{et al.} \cite{ummenhofer2016demon} proposed an end-to-end visual odometry and depth estimation network by formulating structure from motion as a supervised learning problem. However, the work is highly supervised: not only does it require depth and camera motion ground truths, in addition the surface normals and optical flow between images are also required. 

\paragraph{Unsupervised or semi-supervised methods}
Recent works suggest that unsupervised pipeline for learning depth is possible from stereo image pairs using a photometric warp loss to replace a loss based on ground truth depth. Garg \textit{et al.} \cite{garg2016depth} used binocular stereo pairs (for which the inter-camera transformation is known) and trained a network to predict the depth that minimises the photometric difference between the true right image and one synthesized by warping the left image into the right's viewpoint, using the predicted depth. Godard \textit{et al.} \cite{godard2016depth} made improvements to the depth estimation by introducing a symmetric left-right consistency criterion and better stereo loss function. \cite{kuznietsov2017semi} proposed a semi-supervised learning framework by using both sparse depth maps for supervised learning and dense photometric error for unsupervised learning.

An obvious extension to the above framework is to use structure-from-motion techniques to estimate the inter-frame motion (optic flow) \cite{yu2016B2B} instead of depth using the known stereo geometry. But in fact it is possible to go further and to use deep networks also to estimate the camera ego-motion, as shown very recently by \cite{zhou2017sfmlearner} and \cite{vijayanarasimhan2017sfmnet}, both of which use a photometric error for supervising a monocular depth and ego-motion estimation system. 
Similar to other monocular frameworks, \cite{zhou2017sfmlearner} and \cite{vijayanarasimhan2017sfmnet} suffer from scaling ambiguity issue.

Like \cite{garg2016depth}\cite{godard2016depth}, in our work we use stereo pairs for training, since this avoids issues with the depth-speed ambiguity that exist in monocular 3D reconstruction. In addition we jointly train a network to also estimate ego-motion from a pair of images. This allows us to enforce both the temporal {\em and} stereo constraints to improve our depth estimation in a joint framework. 

All of the unsupervised depth estimation methods rely on photo-consistency assumption which gets violated often in practice. To cope with that \cite{garg2016depth}\cite{zhou2017sfmlearner} use robust norms like L1 norm of the warp error.
\cite{godard2016depth} uses hand crafted features like SSIM \cite{wang2004image}. Other handcrafted features like SIFT \cite{lowe2004sift}, HOG \cite{dalal2005hog}, ORB \cite{rublee2011orb} are all usable and can be explored in unsupervised learning framework for robust warping loss. 
More interestingly, one can learn good features specifically for the task of matching.
LIFT \cite{yi2016lift} and MC-CNN \cite{zbontar2016mccnn} learn a similarity measure on small image patches while \cite{weerasekera2017feature}\cite{choy2016ucn} learns fully convolutional features good for matching. 
In our work, we compare the following features for their potential for robust warp error minimization: standard RGB photo-consistency; ImageNet features (conv1); features from \cite{weerasekera2017feature}; features from a ``self-supervised" version of \cite{weerasekera2017feature}; and features derived from our depth network.

\section{Method} \label{sec:method}

\begin{figure*}[h] 
\centering
    \includegraphics[width=1\textwidth]{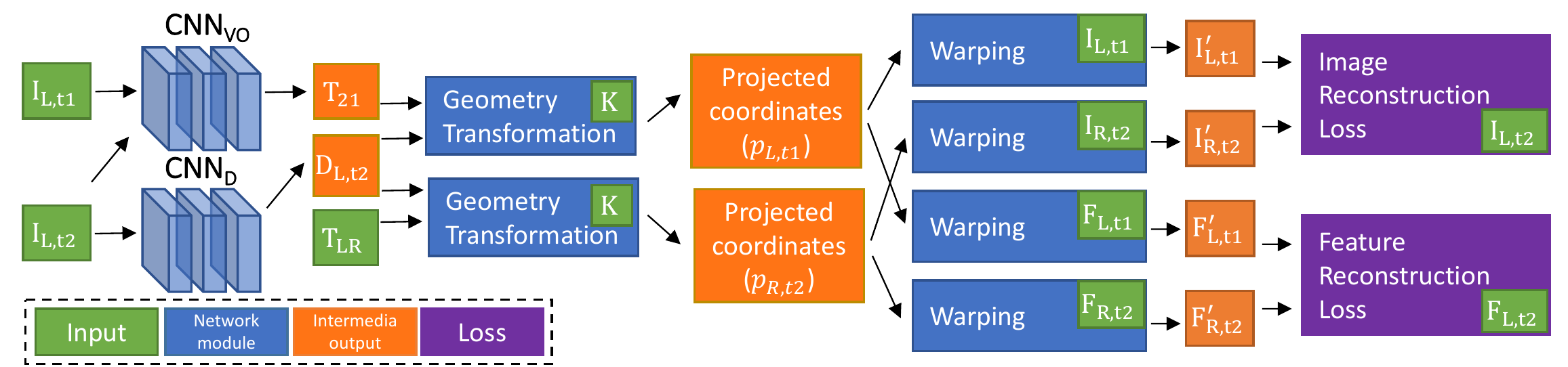}
    \caption{Illustration of our proposed framework in training phase. $\text{CNN}_{VO}$ and $\text{CNN}_D$ can be used independently in testing phase. }\label{fig:framework}
\end{figure*}

This section describes our framework (shown in Figure \ref{fig:framework}) for jointly learning a single view depth ConvNet ($\text{CNN}_{D}$) and a visual odometry ConvNet ($\text{CNN}_{VO}$) from stereo sequences. The stereo sequences learning framework overcomes the scaling ambiguity issue with monocular sequences, and enables the system to take advantage of both left-right (spatial) and forward-backward (temporal) consistency checks. 

\subsection{Image reconstruction as supervision} \label{sec:method_imgSuper}
The fundamental supervision signal in our framework comes from the task of image reconstruction. For two nearby views, we are able to reconstruct the reference view from the live view, given that the depth of the reference view and relative camera pose between two views are known. 
Since the depth and relative camera pose can be estimated by a ConvNet, the inconsistency between the real and the reconstructed view allows the training of the ConvNet. 
However, a monocular framework without extra constraints \cite{zhou2017sfmlearner} suffers from the scaling ambiguity issue. Therefore, we propose a stereo framework which constrains the scene depth and relative camera motion to be in a common, real-world scale, given an extra constraint set by the known stereo baseline.

In our proposed framework using stereo sequences, for each training instance, we have a temporal pair ($I_{L,t1}$ and $I_{L,t2}$) and a stereo pair ($I_{L,t2}$ and $I_{R,t2}$), where $I_{L,t2}$ is the reference view while $I_{L,t1}$ and $I_{R,t2}$ are the live views. We can synthesize two reference views, $I_{L,t1}'$ and $I_{R,t2}'$, from $I_{L,t1}$ and $I_{R,t2}$, respectively. The synthesis process can be represented by, 
\begin{align}
    I_{L,t1}' = f(I_{L,t1}, K, T_{t2 \rightarrow t1}, D_{L,t2}) \label{eqn:synImg1} \\
    I_{R,t2}' = f(I_{R,t2}, K, T_{L \rightarrow R}, D_{L,t2}). \label{eqn:synImg2}
\end{align}
\noindent where $f(.)$ is a synthesis function defined in Sec.\ref{sec:method_diffGeoModules}; 
$D_{L,t2}$ denotes the depth map of the reference view; 
$T_{L \rightarrow R}$ and $T_{t2 \rightarrow t1}$ are the relative camera pose transformations between the reference view and the live views; 
and $K$ denotes the known camera intrinsic matrix. Note that $D_{L,t2}$ is mapped from $I_{L,t2}$ via $\text{CNN}_{D}$ 
while $T_{t2 \rightarrow t1}$ is mapped from $[I_{L,t1}, I_{L,t2}]$ via $\text{CNN}_{VO}$. 

The image reconstruction loss between the synthesized views and the real views are computed as a supervision signal to train $\text{CNN}_{D}$ and $\text{CNN}_{VO}$. The image construction loss is represented by,
\begin{equation}
    L_{ir} = \sum_{p} \left( {|I_{L,t2}(p) - I_{L,t1}'(p)|}  + 
    {|I_{L,t2}(p) - I_{R,t2}'(p)|} \right). \label{eqn:imgLoss}
\end{equation}
The effect of using stereo sequences instead of monocular sequences is two-fold. 
The known relative pose $T_{L \rightarrow R}$ between the stereo pair constrains $\text{CNN}_D$ and $\text{CNN}_{VO}$ to estimate depths and relative pose between the temporal pair in a real-world scale. As a result, our model is able to estimate single view depths and two-view odometry without the scaling ambiguity issue at test time.
Second, in addition to stereo pairs with only one live view, the temporal pair provides a second live view for the reference view. The multi-view scenario takes advantage of the full set of constraints available from the stereo and temporal image pairs.

In this section, we describe an unsupervised framework that learns depth estimation and visual odometry without scaling ambiguity issue using stereo video sequences.

\subsection{Differentiable geometry modules} \label{sec:method_diffGeoModules}

As indicated in Eqn.\ref{eqn:synImg1} - \ref{eqn:synImg2}, an important function in our learning framework is the synthesis function, $f(.)$. The function consists two differentiable operations which allow gradient propagation for the training of the ConvNet. 
The two operations are epipolar geometry transformation and warping. The former defines the correspondence between pixels in two views while the latter synthesize an image by warping a live view.

Let $p_{L,t2}$ be the homogeneous coordinates of a pixel in the reference view. 
We can obtain $p_{L,t2}$'s projected coordinates onto the live views using epipolar geometry, similar to \cite{handa2016gvnn,zhou2017sfmlearner}. The projected coordinates are obtained by
\begin{align}
    p_{R,t2} &= K T_{L \rightarrow R} D_{L,t2}(p_{L,t2}) K^{-1} p_{L,t2} \label{eqn:projCoorR}\\
    p_{L,t1} &= K T_{t2 \rightarrow t1} D_{L,t2}(p_{L,t2}) K^{-1} p_{L,t2} \label{eqn:projCoort1}, 
\end{align}
where $p_{R,t2}$ and $p_{L,t1}$ are the projected coordinates on $I_{R,t2}$ and $I_{L,t1}$ respectively.
Note that $D_{L,t2}(p_{L,t2})$ is the depth at position $p_{L,t2}$;
$T \in SE3$ is a 4x4 transformation matrix defined by 6 parameters, in which a 3D vector $\mathbf{u} \in so3$ is an axis-angle representation and a 3D vector $\mathbf{v} \in \mathbb{R}^{3}$ represents translations.

After getting the projected coordinates from Eqn.\ref{eqn:projCoorR} - \ref{eqn:projCoort1}, new reference frames can be synthesized from the live frames using the differentiable bilinear interpolation mechanism (warping) proposed in \cite{jaderberg2015stn}. 

\subsection{Feature reconstruction as supervision}\label{sec:method_featSuper}
The stereo framework we proposed above implicitly assumes that the scene is Lambertian, so that the brightness is constant regardless the observer's angle of view. This condition implies that the image reconstruction loss is meaningful for training the ConvNets.
Any violation of the assumption can potentially corrupt the training process by propagating the wrong gradient back to the ConvNets. 
To improve the robustness of our framework, we propose a feature reconstruction loss: 
instead of using 3-channel color intensity information solely (image reconstruction loss), we explore the use of dense features as an additional supervision signal.

Let $F_{L,t2}$, $F_{L,t1}$ and $F_{R,t2}$ be the corresponding dense feature representations of $I_{L,t2}$, $I_{L,t1}$ and $I_{R,t2}$ respectively. 
Similar to the image synthesis process, two reference views, $F_{L,t1}'$ and $F_{R,t2}'$, can be synthesized from $F_{L,t1}$ and $F_{R,t2}$, respectively.
The synthesis process can be represented by, 
\begin{align}
    F_{L,t1}' = f(F_{L,t1}, K, T_{t2 \rightarrow t1}, D_{L,t2}) \label{eqn:synFeat1} \\
    F_{R,t2}' = f(F_{R,t2}, K, T_{L \rightarrow R}, D_{L,t2}).
    \label{eqn:synFeat2}
\end{align}
Then, the feature reconstruction loss can be formulated as,
\begin{equation} 
    L_{fr} = \sum_{p}{|F_{L,t2}(p) - F_{L,t1}'(p)|} 
    +\sum_{p}{|F_{L,t2}(p) - F_{R,t2}'(p)|} \label{eqn:featLoss} 
\end{equation}

In this work, we explore four possible dense features, as detailed in Section \ref{sec:ablation}.

\subsection{Training loss} \label{sec:method_loss}

As introduced in Sec.\ref{sec:method_imgSuper} and Sec.\ref{sec:method_featSuper}, the main supervision signal in our framework comes from the image reconstruction loss while the feature reconstruction loss acts as an auxiliary supervision. Furthermore, similar to \cite{garg2016depth}\cite{zhou2017sfmlearner}\cite{godard2016depth}, we have a depth smoothness loss which encourages the predicted depth to be smooth.

To obtain a smooth depth prediction, following the approach adopted by \cite{heise2013pm}\cite{godard2016depth}, we encourage depth to be smooth locally by introducing an edge-aware smoothness term. The depth discontinuity is penalized if image continuity is showed in the same region. Otherwise, the penalty is small for discontinued depths.
The edge-aware smoothness loss is formulate as
\begin{equation}
    L_{ds} = \sum_{m,n}^{W,H}
    |\partial_x D_{m,n}| e^{-|{\partial_x I_{m,n}}|}+
    |\partial_y D_{m,n}| e^{-|{\partial_y I_{m,n}}|},
\end{equation}
where $\partial_x(.)$ and $\partial_y(.)$ are gradients in horizontal and vertical direction respectively. Note the $D_{m,n}$ is inverse depth in the above regularization.

The final loss function becomes 
\begin{equation}
    L =
    \lambda_{ir} L_{ir} + 
    \lambda_{fr} L_{fr} +
    \lambda_{ds} L_{ds},
\end{equation}
where $\lambda_{ir}$, $\lambda_{fr}$ and $\lambda_{ds}$ are the loss weightings for each loss term.

\subsection{Network architecture} \label{sec:method_netArchi}
\paragraph{Depth estimation}

Our depth ConvNet is composed of two parts, encoder and decoder. 
For the encoder, we adopt the convolutional network in a variant of ResNet50 \cite{he2016resnet} with half filters (ResNet50-1by2)  for the sake of computation cost. The ResNet50-1by2 contains less than 7 million parameters which is around one fourth of the original ResNet50. For the decoder network, the decoder firstly converts the encoder output (1024-channel feature maps) into a single channel feature map using a 1x1 kernel, followed by conventional bilinear upsampling kernels with skip-connections.
Similar to \cite{long2015fcn}\cite{garg2016depth}\cite{godard2016depth}, the decoder uses skip-connections to fuse low-level features from different stages of the encoder. 
We use ReLU activation after the last prediction layer to ensure positive prediction comes from the depth ConvNet. For the output of the depth ConvNet, we design our framework to predict inverse depth instead of depth. However, the ReLU activation may cause zero estimation which results in infinite depth. Therefore, we convert the predicted inverse depth to depth by $D = 1/(D_{inv}+10^{-4})$.

\paragraph{Visual odometry}

The visual odometry ConvNet is designed to take two concatenated views along the color channels as input and output a 6D vector $[\mathbf{u},\mathbf{v}] \in se3$, which is then converted to a 4x4 transformation matrix. 
The network is composed of 6 stride-2 convolutions followed by 3 fully-connected layers. The last fully-connected layer gives the 6D vector, which defines the transformation from reference view to live view $T_{ref \rightarrow live}$. 

\section{Experiments} \label{sec:expRes}
In this section we show extensive experiments for evaluating the performance of our proposed framework. We favorably compare our approach on KITTI dataset \cite{Geiger2012kitti}\cite{Geiger2013kitti} with prior art on both single view depth and visual odometry estimation. 
Additionally, we perform a detailed ablation study on our framework to show that using temporal consistency while training and use of learned deep features along with color consistency both improves the single view depth predictions.
Finally, we show two variants of deep features and the corresponding effect, which we show examples of using deep features for dense matching.

We train all our CNNs with the Caffe \cite{jia2014caffe} framework. 
We use Adam optimizer with the proposed optimization settings in \cite{kingma2014adam} with 
$[\beta_1, \beta_2, \epsilon] = [0.9, 0.999, 10^{-8}]$. The initial learning rate is 0.001 for all the trained network, which we decrease manually when the training loss converges. For the loss weighting in our final loss function, we empirically find that the combination $[\lambda_{ir}, \lambda_{fr}, \lambda_{ds}] = [1, 0.1, 10]$ results in a stable training. No data augmentation is involved in our work.

Our system is trained mainly in KITTI dataset \cite{Geiger2013kitti}\cite{Geiger2012kitti}. 
The dataset contains 61 video sequences with 42,382 rectified stereo pairs, with the original image size being 1242x375 pixels. However, we use image size of 608x160 in our training setup for the sake of computation cost. 
We use two different splits of the KITTI dataset for evaluating estimated ego-motion and depth. For single view depth estimation, we follow the Eigen split provided by \cite{eigen2014depth} for fair comparisons with \cite{garg2016depth,godard2016depth,eigen2014depth,liu2016depth}. 
On the other hand, in order to evaluate our visual odometry performance and compare to prior approaches, we follow \cite{zhou2017sfmlearner} by training both the depth and pose network on the official KITTI Odometry training set. Note that there are overlapping scenes between two splits (i.e. some testing scenes of Eigen Split are included in the training scenes of Odometry Split, and vice versa). Therefore, finetuning/testing models trained in any split to another split is not allowable/sensible.
The detail about both splits are:

\noindent\textbf{Eigen Split} 
Eigen \textit{et al.} \cite{eigen2014depth} select 697 images from 28 sequences as test set for single view depth evaluation. The remaining 33 scenes contains 23,488 stereo pairs for training. We  follow this setup and form 23,455 temporal stereo pairs. 

\noindent\textbf{Odometry Split}
The KITTI Odometry Split \cite{Geiger2012kitti} contains 11 driving sequences with publicly available ground truth camera poses.
% obtained through the IMU/GPS readings. 
We follow \cite{zhou2017sfmlearner} to train our system on the Odometry Split (no finetuning from Eigen Split is performed). The split in which sequences 00-08 (sequence 03 is not available in KITTI Raw Data) are used for training while 09-10 are used for evaluation. The training set contains 8 sequences with 19,600 temporal stereo pairs.

For each dataset split, we form temporal pairs by choosing frame $I_t$ as the live frame while frame $I_{t+1}$ as the reference frame -- to which the live frame is warped. The reason for this choice is that as the mounted camera in KITTI moves forward, most pixels in $I_{t+1}$ have correspondence in $I_{t}$ giving us a better warping error.

\subsection{Visual odometry results}\label{sec:expRes_VO}
We use the Odometry Split mentioned above to evaluate the performance of our frame to frame odometry estimation network.
The result is compared with the monocular training based network \cite{zhou2017sfmlearner} and a popular SLAM system -- ORB-SLAM \cite{mur2015orbslam} (with and without loop closure) as very strong baselines. 
Both of the ORB-SLAM versions use local bundle adjustment and more importantly a single scale map to assist the tracking. We ignore the frames (First 9 and 30 respectively) from the sequences (09 and 10) for which ORB-SLAM fails to bootstrap with reliable camera poses due to lack of good features and large rotations. Following the KITTI Visual Odometry dataset evaluation criterion we use possible sub-sequences of length (100, 200, ... , 800) meters and report the average translational and rotational errors for the testing
% validation: 
sequence 09 and 10 in Table \ref{table:vo_benchmark}.

\begin{table} [t] 
\begin{center}
\resizebox{1\columnwidth}{!}{%
\begin{tabular}{| l ||  c c | c c | 
% c c |
}
\hline
Method &  
\multicolumn{2}{c|}{Seq. 09} &
\multicolumn{2}{c|}{Seq. 10} 
\\
 & $t_{err} (\%)$ & $r_{err}(^\circ/100m)$ & 
 $t_{err} (\%)$ & $r_{err}(^\circ/100m)$ 
\\
\hline%\hline
ORB-SLAM (LC) \cite{mur2015orbslam} & 
16.23 & 1.36 &
 / & / 
\\

ORB-SLAM \cite{mur2015orbslam} & 
15.30 & 0.26 &
3.68 & 0.48
 \\

Zhou \textit{et al.}\cite{zhou2017sfmlearner} & 
17.84 & 6.78 &
 37.91 & 17.78  
\\

\textbf{Ours (Temporal)} &
11.93 & 3.91 &
 12.45 & 3.46  
\\

\textbf{Ours (Full-NYUv2)} &
11.92 & 3.60  &
12.62 & 3.43  
 \\
 
\hline
\end{tabular}
}
\end{center}
\caption{Visual odometry result evaluated on Sequence 09, 10 of KITTI Odometry dataset. $t_{err}$ is average translational drift error. $r_{err}$ is average rotational drift error.}
\label{table:vo_benchmark}
\end{table}

\begin{figure}[t] 
\centering
    \includegraphics[width=1\columnwidth]{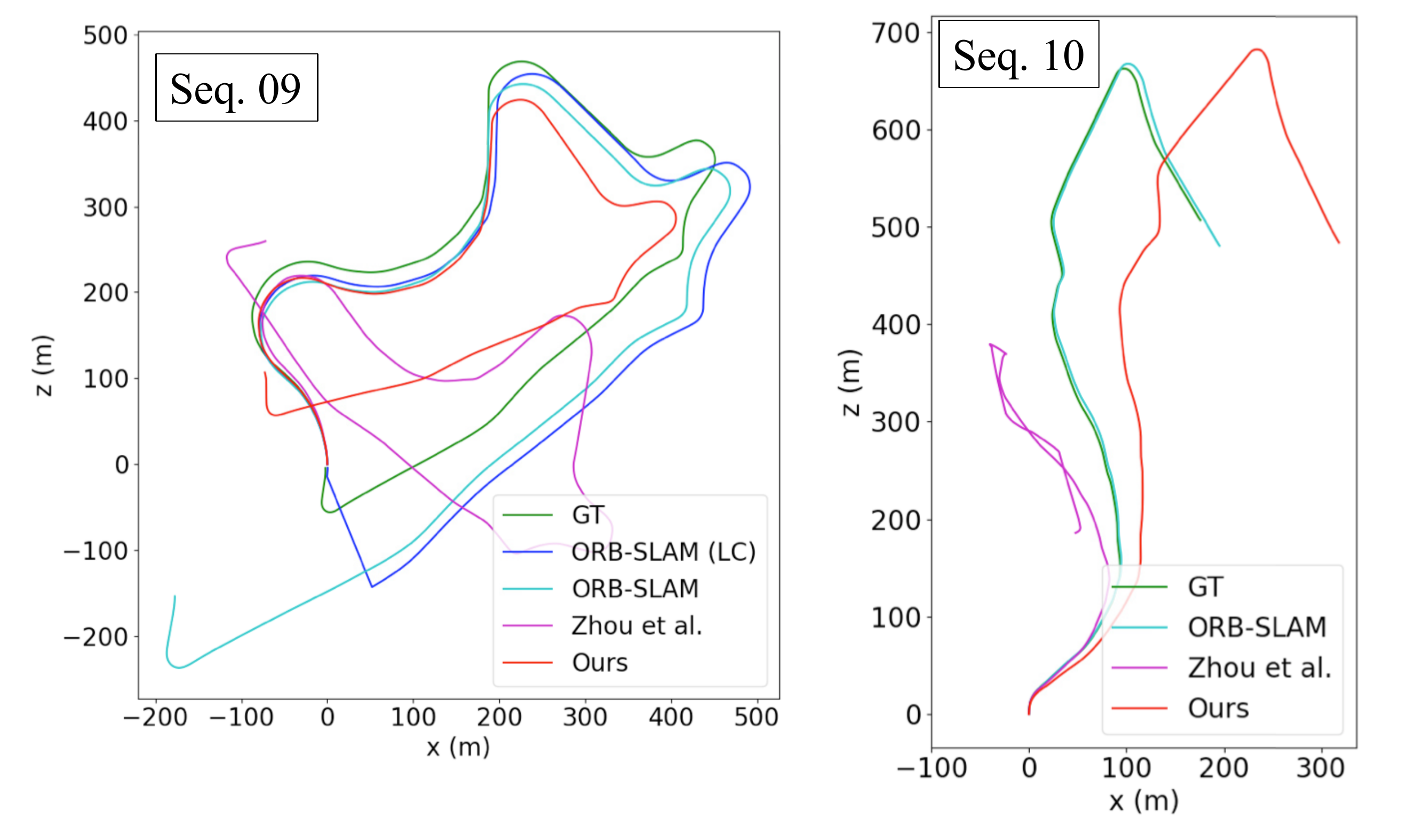}
    \caption{Qualitative result on visual odometry. Full trajectories on the testing sequences (09, 10) are plotted. }\label{fig:voTrajEg}
\end{figure}

\begin{figure}[t] 
\centering
    \includegraphics[width=1\columnwidth]{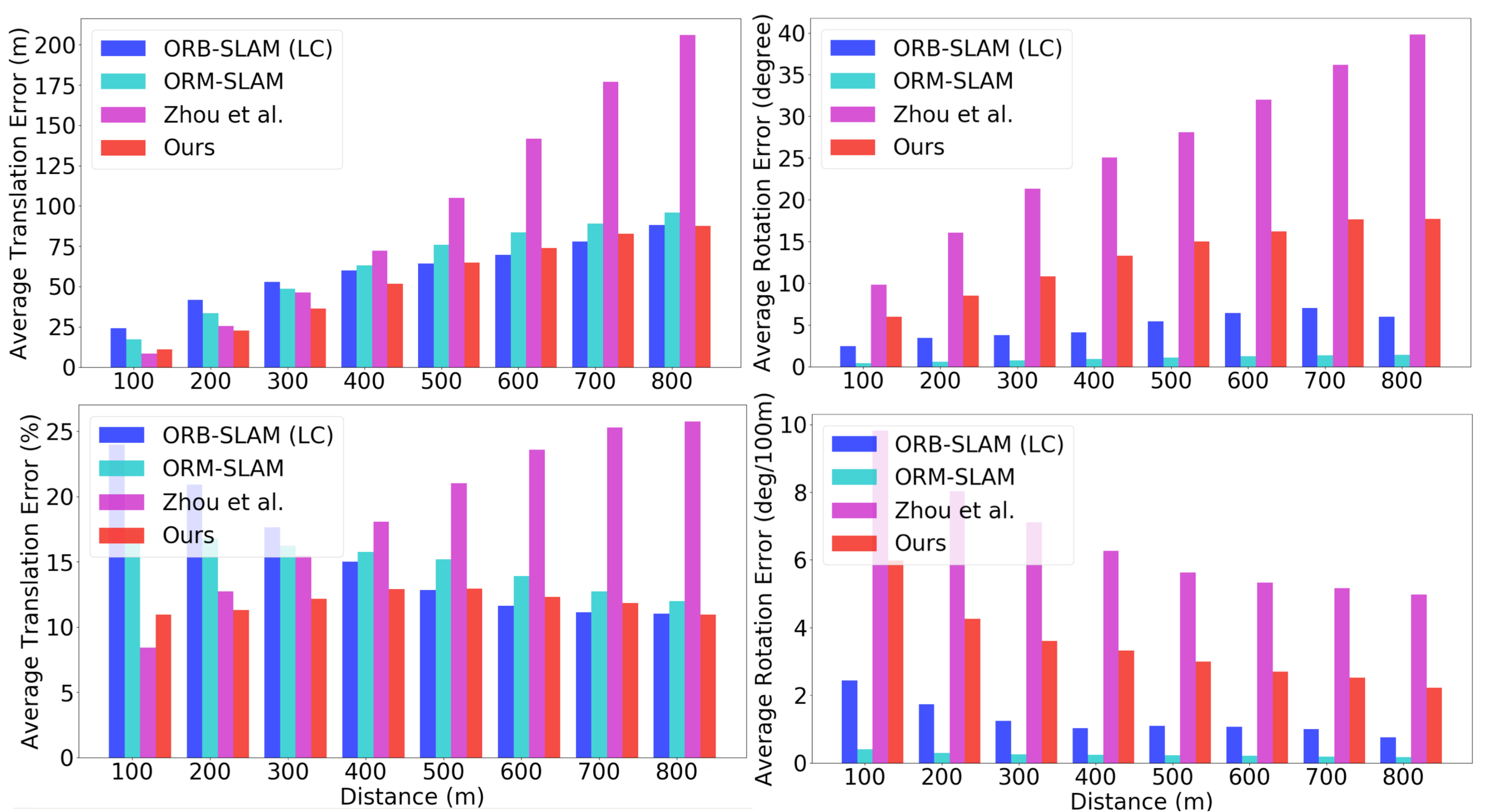}
    \caption{Comparison of VO error with different translation threshold for sequence 09 of odometry dataset.}
    \label{fig:voErrAnalysis09}
\end{figure}

As ORB-SLAM suffers from a single depth-translation scale ambiguity for the whole sequence, we align the ORB-SLAM trajectory with ground-truth by optimizing the map scale following standard protocol. 
For our method, we simply integrate the estimated frame-to-frame camera poses over the entire sequence without any post processing. 
Frame-to-frame pose estimation of \cite{zhou2017sfmlearner} only avails small 5-frame long tracklets, each of which is already aligned independently with the ground-truth by fixing translation scales. This translation normalization leaves \cite{zhou2017sfmlearner}'s error to only indicate the relative translation magnitudes error over small sequences. 
As the KITTI sequences are recorded by camera mounted on a car which mostly move forward, even average 6DOF motion as reported in \cite{zhou2017sfmlearner} overperforms frame-to-frame odometry methods (ORB-SLAM when used only on 5 frames does not bootstrap mapping). Nonetheless we simply integrate the aligned tracklets to estimate the full trajectory for \cite{zhou2017sfmlearner} and evaluate. 
It is important to note that this evaluation protocol is highly disadvantageous to the proposed method as no scope for correcting the drift or translation scale is permitted. A visual comparison of the estimated trajectories for all the methods can be seen in Figure \ref{fig:voTrajEg}.

As can be seen in Table \ref{table:vo_benchmark}, our stereo based odometry learning method outperforms monocular learning method \cite{zhou2017sfmlearner} by a large margin even without any further post-processing to fix translation scales. Our method is able to give comparable odometry results on sequence 09 to that of the full ORB-SLAM and respectable trajectory for sequence 10 on which larger error in our frame to frame rotation estimation leads to a much larger gradual drift which should be fixed by bundle adjustment. 

To further compare the effect of bundle adjustment, we evaluate the average errors for different translation bins  and report the result for sequence 09 in Figure \ref{fig:voErrAnalysis09}.
It can be seen clearly that both our method and \cite{zhou2017sfmlearner} are better than ORB-SLAM when the translation magnitude is small. As translation magnitude increases, the simple integration of frame to frame VO starts drifting gradually, which suggests a clear advantage of a map based tracking over frame to frame VO without bundle adjustment.

\subsection{Depth estimation results}

\begin{table*} [t] 
\begin{center}
\resizebox{2\columnwidth}{!}{%
\begin{tabular}{| l c c || c c c c | c c c|}
\hline
Method & Dataset & Supervision & 
\multicolumn{4}{c|}{Error metric} &
\multicolumn{3}{c|}{Accuracy metric} \\
 & & & Abs Rel & SqRel & RMSE & RMSE log &
% Abs Rel & Sq Rel & RMSE & RMSE log &
$\delta<1.25$ & $\delta<1.25^2$ & $\delta<1.25^3$
\\
\hline\hline
\multicolumn{10}{|c|}{Depth: cap 80m} \\
\hline
Train set mean & K & Depth & 
0.361 & 4.826 & 8.102 & 0.377 &
0.638 & 0.804 & 0.894\\

Eigen \textit{et al.} \cite{eigen2014depth} Fine & K & Depth & 
0.203 & 1.548 & 6.307 & 0.282 &
0.702 & 0.890 & 0.958\\

Liu \textit{et al.} \cite{liu2016depth} & K & Depth & 
0.201 & 1.584 & 6.471 & 0.273 &
0.680 & 0.898 & 0.967\\

Zhou \textit{et al.} \cite{zhou2017sfmlearner}  & K & Mono. & 
0.208 & 1.768 & 6.856 & 0.283 &
0.678 & 0.885 & 0.957\\

Garg \textit{et al.} \cite{garg2016depth}  & 
K & Stereo & 
0.152 & 1.226 & 5.849 & 0.246 & 
0.784 & 0.921 & 0.967  \\

Godard \textit{et al.} \cite{godard2016depth} & K & Stereo & 
0.148 & 1.344 & 5.927 & 0.247 &
0.803 & 0.922 & 0.964\\

\textbf{Ours} (Temporal) & 
K & Stereo & 
0.144 & 1.391 & 5.869 & 0.241 &
0.803 & 0.928 & 0.969 \\

\textbf{Ours} (Full-NYUv2) & 
K & Stereo & 
0.135 & 1.132 & 5.585 & 0.229 &
0.820 & 0.933 & 0.971 \\

 \hline\hline
\multicolumn{10}{|c|}{Depth: cap 50m} \\
\hline

Zhou \textit{et al.} \cite{zhou2017sfmlearner}  & 
K & Mono. & 
0.201 & 1.391 & 5.181 & 0.264 &
0.696 & 0.900 & 0.966\\

Garg \textit{et al.} \cite{garg2016depth}  & 
K & Stereo & 
0.169 & 1.080 & 5.104 & 0.273 &
0.740 & 0.904 & 0.962\\

Godard \textit{et al.} \cite{godard2016depth} & 
K & Stereo & 
0.140 & 0.976 & 4.471 & 0.232 &
0.818 & 0.931 & 0.969\\

\textbf{Ours} (Temporal) & 
K & Stereo &
0.135 & 0.905 & 4.366 & 0.225 &
0.818 & 0.937 & 0.973\\

\textbf{Ours} (Full-NYUv2) & 
K & Stereo & 
0.128 & 0.815 & 4.204 & 0.216 &
0.835 & 0.941 & 0.975\\

\hline
\end{tabular}
}
\end{center}
\caption{Comparison of single view depth estimation performance with existing approaches. For training, K is KITTI dataset (Eigen Split). For a fair comparison, all methods (except \cite{eigen2014depth}) are evaluated on the cropped region from \cite{godard2016depth}. For the supervision, ``Depth" means ground truth depth is used in the method; ``Mono." means monocular sequences are used in the training; ``Stereo" means stereo sequences with known stereo camera poses in the training. 
}
\label{table:depth_benchmark}
\end{table*}

% UPDATEME
\begin{figure*}[!t] 
\centering
\includegraphics[width=2.1\columnwidth]{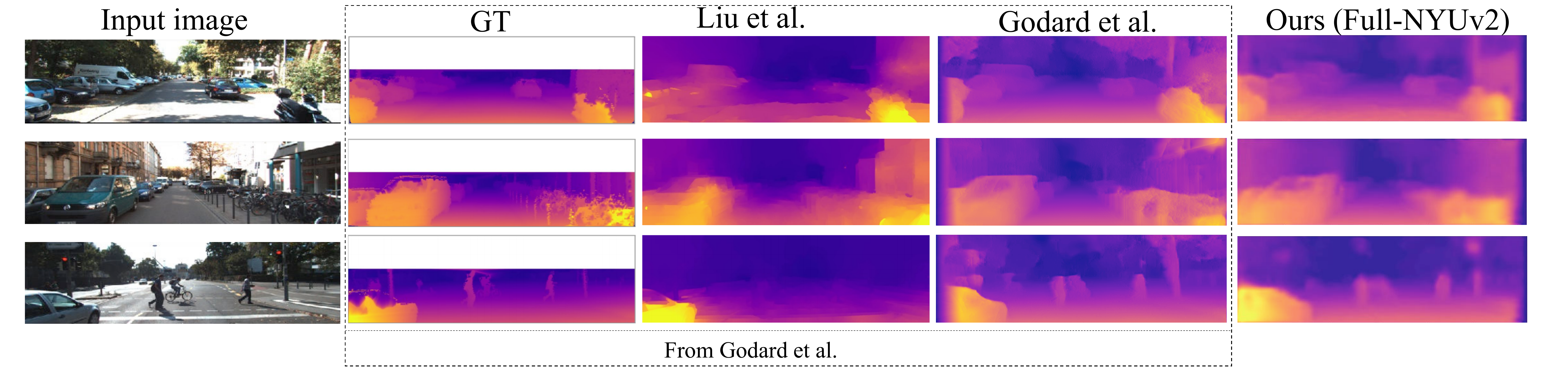}
\caption{Single view depth estimation examples in Eigen Split. The ground truth depth is interpolated for visualization purpose. }\label{fig:depthEg}
\vspace{-5mm}
\end{figure*}

\begin{table*} [t] 
\begin{center}
% \vspace{-5mm}
\resizebox{2\columnwidth}{!}{%
\begin{tabular}{| l || c c c || c c c c| c c c|}
\hline
Method &  
Stereo & Temporal & Feature & 
\multicolumn{4}{c|}{Error metric} &
\multicolumn{3}{c|}{Accuracy metric} \\
 & & & & Abs Rel & SqRel & RMSE & RMSE log &
% Abs Rel & Sq Rel & RMSE & RMSE log &
$\delta<1.25$ & $\delta<1.25^2$ & $\delta<1.25^3$
\\
\hline\hline

\multicolumn{11}{|c|}{Encoder: ResNet50-1by2; Decoder: Bilinear upsampler} \\
\hline

Baseline & 
\cmark & \xmark & \xmark &
0.143 & 0.859 & 4.310 & 0.229 &
0.802 & 0.933 & 0.973\\

Temporal & 
\cmark & \cmark & \xmark &
0.135 & 0.905 & 4.366 & 0.225 &
0.818 & 0.937 & 0.973\\

ImageNet Feat. & 
\cmark & \xmark & \cmark &
0.136 & 0.880 & 4.390 & 0.230 &
0.823 & 0.935 & 0.970\\

KITTI Feat. & 
\cmark & \xmark & \cmark &
0.130 & 0.860 & 4.271 & 0.221 &
0.831 & 0.938 & 0.973\\

NYUv2 Feat. & 
\cmark & \xmark & \cmark &
0.132 & 0.906 & 4.279 & 0.220 &
0.831 & 0.939 & 0.974\\

Full-NYUv2 &
\cmark & \cmark & \cmark &
0.128 & 0.815 & 4.204 & 0.216 &
0.835 & 0.941 & 0.975\\

\hline \hline

\multicolumn{11}{|c|}{Encoder: ResNet50-1by2; Decoder: Learnable upsampler}\\
\hline

Baseline2 & 
\cmark & \xmark & \xmark &
0.155 & 1.307 & 4.560 & 0.242 &
0.805 & 0.928 & 0.968\\

Temporal2 & 
\cmark & \cmark & \xmark &
0.141 & 0.998 & 4.354 & 0.232 &
0.814 & 0.932 & 0.971 \\

Depth Feat. & 
\cmark & \xmark & \cmark &
0.142 & 0.956 & 4.377 & 0.230 &
0.817 & 0.934 & 0.971\\

Full-Depth & 
\cmark & \cmark & \cmark &
0.137 & 0.893 & 4.348 & 0.228 &
0.821 & 0.935 & 0.971 \\

\hline

\end{tabular}
}
\end{center}
\caption{Ablation study on single view depth estimation. The result is evaluated in KITTI 2015 using Eigen Split test set, following the evaluation protocol proposed in \cite{godard2016depth}. The results are capped at 50m depth. Stereo: stereo pairs are used for training; Temporal: additional temporal pairs are used; Feature: feature reconstruction loss is used.
\vspace{-5mm}
}
\label{table:depth_ablation}
\end{table*}

\begin{figure} 
\centering
    \includegraphics[width=1\columnwidth]{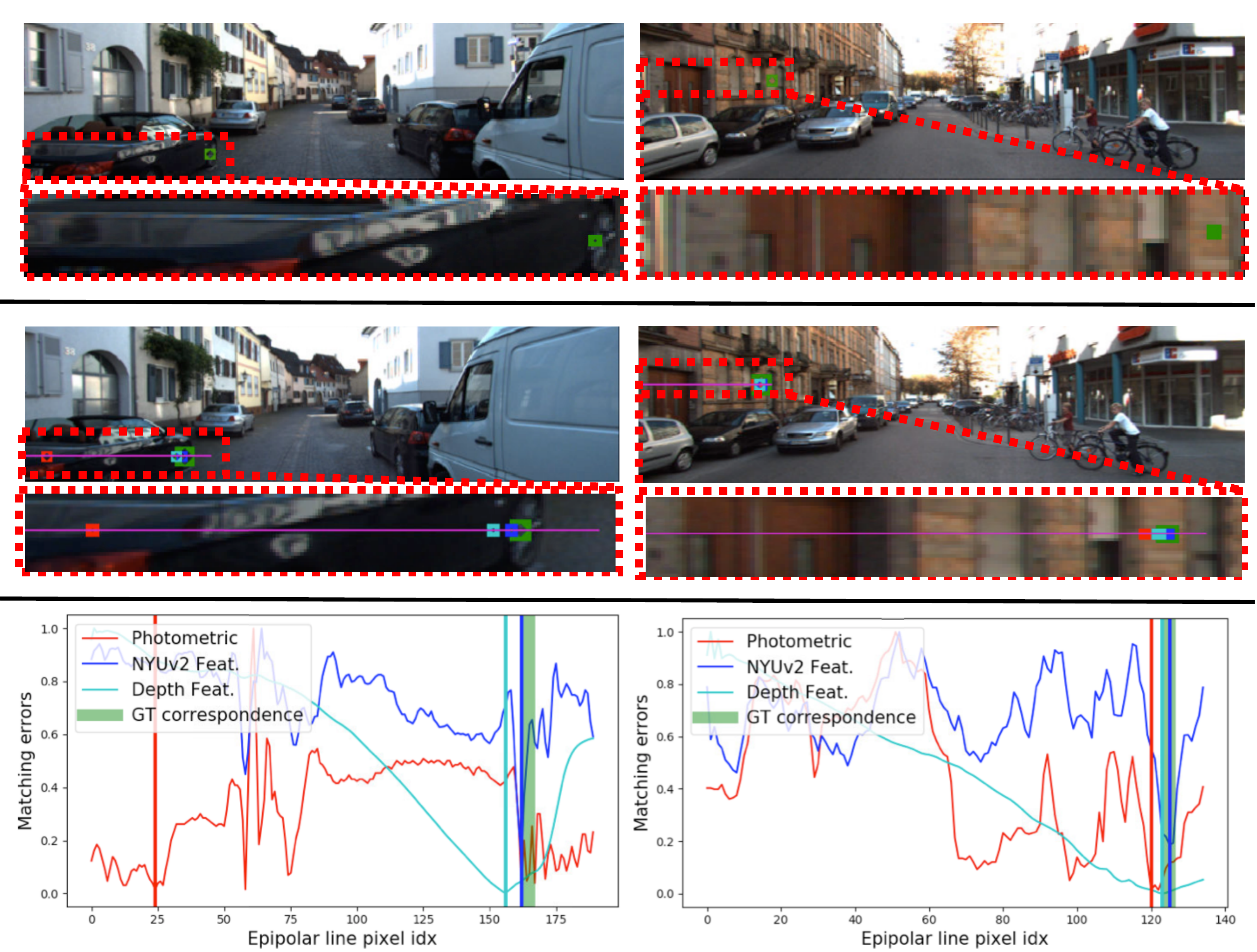}
    \caption{Stereo matching examples. Rows: (1) Left image; (2) Right image; (3) Matching error using color intensity and deep features. Photometric loss is not robust when compared with feature loss, especially in ambiguous regions.}\label{fig:stereoMatchEg}
\end{figure}

We use the Eigen Split to evaluate our system and compare the results with various state of the art depth estimation methods. Following the evaluation protocol proposed in \cite{godard2016depth} which uses the same crop as \cite{garg2016depth}, 
we use both the 50m and 80m threshold of maximum depth for evaluation and report all standard error measures in Table \ref{table:depth_benchmark} with some visual examples in Figure \ref{fig:depthEg}.
As shown in \cite{garg2016depth}, photometric stereo based training with AlexNet-FCN architecture and Horn and Schunck \cite{horn1981determining} loss already gave more accurate results than the state of the art supervised methods \cite{eigen2014depth}\cite{liu2016depth} on KITTI. 
For fair comparison of \cite{garg2016depth} with other methods we evaluate the results reported by the authors publicly with 80m cap on maximum depth. 
All methods using stereo for training are substantially better than \cite{zhou2017sfmlearner} which is using only monocular training. 
Benefited by the feature based reconstruction loss and  additional warp error via odometry network, our method outperforms both \cite{garg2016depth} and \cite{godard2016depth} with reasonable margin. It is important to note that unlike \cite{godard2016depth} left-right consistency, data augmentation, run-time shuffle, robust similarity measure like SSIM\cite{wang2004image} are not used to train our network and should lead to further improvement.

\subsection{Ablation studies}
\label{sec:ablation}

Table \ref{table:depth_ablation} shows an ablation study on depth estimation for our method showing importance of each component of the loss function. 
Our first baseline is a simple architecture (ResNet50-1by2 as encoder; Bilinear upsampler as decoder)
% with relu activation 
trained on the stereo pairs with the loss described in Sec.\ref{sec:method_loss} which closely follows \cite{garg2016depth} (GitHub version). 
When we train the pose network jointly with the depth network, we get a slight improvement in depth estimation accuracy. 
Using features from ImageNet feature (conv1 features from pretrained ResNet50-1by-2) improves depth estimation accuracy slightly.
In addition, using features from an off-the-shelf image descriptor \cite{weerasekera2017feature} gives a further boost. However, \cite{weerasekera2017feature} is trained using NYUv2 dataset \cite{Silberman2012nyuv2} (ground truth poses and depths are required) so we follow \cite{weerasekera2017feature} to train an image descriptor using KITTI dataset but using the estimated poses and depths generated from Method ``Temporal" as pseudo ground truths. Using the features extracted from the self-supervised descriptor (KITTI Feat.) gives a comparable result with that of \cite{weerasekera2017feature}. 
The system having all three components (Stereo + Temporal + NYUv2 Feat.) performs best as can be seen in the top part of Table \ref{table:depth_ablation}.  

As most other unsupervised depth estimation methods use a convolutional encoder with deconvnet architecture like \cite{noh2015learning}\cite{ronneberger2015u} for dense predictions, we also experimented with learnable deconv architecture with the ResNet50-1by2 as encoder -- learnable upsampler as decoder setup. The results in the bottom part of the table reflects that overall performance of this Baseline2 was slightly inferior to the first baseline. 
To improve the performance of this baseline, we explore the use of deep features extracted from the depth decoder itself. 
At the end the decoder outputs a 32-channel feature map which we directly use for feature reconstruction loss. Using these self-embedded depth features for additional warp error minimization also shows promising improvements in the accuracy of the depth predictions without requiring any explicit supervision for matching as required by \cite{weerasekera2017feature}.

In Figure \ref{fig:stereoMatchEg}, we compare the deep features of \cite{weerasekera2017feature} and the self-embedded depth features against color consistency on the task of stereo matching. Photometric error is not as robust as deep feature error, especially in texture-less regions, there are multiple local minima with similar magnitude. However, both NYUv2 Feature from \cite{weerasekera2017feature} and self-embedded depth features show distinctive local minimum which is a desirable property.

\section{Conclusion} \label{sec_conclusion}
We have presented an unsupervised learning framework for single view depth estimation and monocular visual odometry using stereo data for training. 
We have shown that the use of binocular stereo sequences for jointly learning the two tasks, enable odometry prediction in \emph{metric scale} simply given 2 frames
% We have shown that the use of binocular stereo sequences for learning the two tasks jointly predicts odometry in metric scale using 2 frames only. % in a monocular training framework.
We also show the advantage of using temporal image alignment, in addition to stereo pair alignment for single view depth predictions. % relative to the state of the art. 
Additionally, we have proposed a novel feature reconstruction loss to have state-of-the-art unsupervised single view depth and frame-to-frame odometry without scale ambiguity.  

There are still a number of challenges to be addressed. 
Our framework assumes no occlusion and the scene is assumed to be rigid. Modelling scene dynamics and occlusions explicitly, in a deep learning framework will provide a natural means for more practical and useful navigation in real scenarios. 
Although we show odometry results that are comparable to the best two-frame estimates available the current systems do not compare favourably with state-of-the-art SLAM systems. An extensive study of CNN architectures more suitable for odometry estimation and a possible way of integrating the map information over time are challenging but very fruitful future directions. 

\section{Acknowledgement}
This work was supported by the UoA Scholarship to HZ and KL, the ARC Laureate Fellowship FL130100102 to IR and the Australian Centre of Excellence for Robotic Vision CE140100016.

\clearpage
{\small
\bibliographystyle{ieee}
% \bibliography{egbib}

\begin{thebibliography}{10}\itemsep=-1pt

\bibitem{agrawal2015seebymoving}
P.~Agrawal, J.~Carreira, and J.~Malik.
\newblock Learning to see by moving.
\newblock In {\em Proceedings of the IEEE International Conference on Computer
  Vision}, pages 37--45, 2015.

\bibitem{choy2016ucn}
C.~B. Choy, J.~Gwak, S.~Savarese, and M.~Chandraker.
\newblock Universal correspondence network.
\newblock In {\em Advances in Neural Information Processing Systems}, pages
  2414--2422, 2016.

\bibitem{dalal2005hog}
N.~Dalal and B.~Triggs.
\newblock Histograms of oriented gradients for human detection.
\newblock In {\em Computer Vision and Pattern Recognition, 2005. CVPR 2005.
  IEEE Computer Society Conference on}, volume~1, pages 886--893. IEEE, 2005.

\bibitem{davison2007monoslam}
A.~J. Davison, I.~D. Reid, N.~D. Molton, and O.~Stasse.
\newblock Monoslam: Real-time single camera slam.
\newblock {\em IEEE transactions on pattern analysis and machine intelligence},
  29(6):1052--1067, 2007.

\bibitem{eigen2014depth}
D.~Eigen, C.~Puhrsch, and R.~Fergus.
\newblock Depth map prediction from a single image using a multi-scale deep
  network.
\newblock In {\em Advances in neural information processing systems}, pages
  2366--2374, 2014.

\bibitem{garg2016depth}
R.~Garg, V.~K. B~G, G.~Carneiro, and I.~Reid.
\newblock Unsupervised cnn for single view depth estimation: Geometry to the
  rescue.
\newblock In {\em European Conference on Computer Vision}, pages 740--756.
  Springer, 2016.

\bibitem{Geiger2013kitti}
A.~Geiger, P.~Lenz, C.~Stiller, and R.~Urtasun.
\newblock Vision meets robotics: The kitti dataset.
\newblock {\em International Journal of Robotics Research (IJRR)}, 2013.

\bibitem{Geiger2012kitti}
A.~Geiger, P.~Lenz, and R.~Urtasun.
\newblock Are we ready for autonomous driving? the kitti vision benchmark
  suite.
\newblock In {\em Conference on Computer Vision and Pattern Recognition
  (CVPR)}, 2012.

\bibitem{godard2016depth}
C.~Godard, O.~{Mac Aodha}, and G.~J. Brostow.
\newblock Unsupervised monocular depth estimation with left-right consistency.
\newblock In {\em CVPR}, 2017.

\bibitem{handa2016gvnn}
A.~Handa, M.~Bloesch, V.~P{\u{a}}tr{\u{a}}ucean, S.~Stent, J.~McCormac, and
  A.~Davison.
\newblock gvnn: Neural network library for geometric computer vision.
\newblock In {\em Computer Vision--ECCV 2016 Workshops}, pages 67--82.
  Springer, 2016.

\bibitem{he2016resnet}
K.~He, X.~Zhang, S.~Ren, and J.~Sun.
\newblock Deep residual learning for image recognition.
\newblock In {\em Proceedings of the IEEE conference on computer vision and
  pattern recognition}, pages 770--778, 2016.

\bibitem{heise2013pm}
P.~Heise, S.~Klose, B.~Jensen, and A.~Knoll.
\newblock Pm-huber: Patchmatch with huber regularization for stereo matching.
\newblock In {\em Proceedings of the IEEE International Conference on Computer
  Vision}, pages 2360--2367, 2013.

\bibitem{horn1981determining}
B.~K. Horn and B.~G. Schunck.
\newblock Determining optical flow.
\newblock {\em Artificial intelligence}, 17(1-3):185--203, 1981.

\bibitem{jaderberg2015stn}
M.~Jaderberg, K.~Simonyan, A.~Zisserman, et~al.
\newblock Spatial transformer networks.
\newblock In {\em Advances in Neural Information Processing Systems}, pages
  2017--2025, 2015.

\bibitem{yu2016B2B}
J.~Y. Jason, A.~W. Harley, and K.~G. Derpanis.
\newblock Back to basics: Unsupervised learning of optical flow via brightness
  constancy and motion smoothness.
\newblock In {\em European Conference on Computer Vision}, pages 3--10.
  Springer, 2016.

\bibitem{jia2014caffe}
Y.~Jia, E.~Shelhamer, J.~Donahue, S.~Karayev, J.~Long, R.~Girshick,
  S.~Guadarrama, and T.~Darrell.
\newblock Caffe: Convolutional architecture for fast feature embedding.
\newblock In {\em Proceedings of the 22nd ACM international conference on
  Multimedia}, pages 675--678. ACM, 2014.

\bibitem{kendall2017deepstereo}
A.~Kendall, H.~Martirosyan, S.~Dasgupta, P.~Henry, R.~Kennedy, A.~Bachrach, and
  A.~Bry.
\newblock End-to-end learning of geometry and context for deep stereo
  regression.
\newblock In {\em Proceedings of the International Conference on Computer
  Vision ({ICCV})}, 2017.

\bibitem{kingma2014adam}
D.~Kingma and J.~Ba.
\newblock Adam: A method for stochastic optimization.
\newblock {\em arXiv preprint arXiv:1412.6980}, 2014.

\bibitem{kuznietsov2017semi}
Y.~Kuznietsov, J.~St{\"u}ckler, and B.~Leibe.
\newblock Semi-supervised deep learning for monocular depth map prediction.
\newblock In {\em Computer Vision and Pattern Recognition (CVPR), 2017 IEEE
  Conference on}, pages 2215--2223. IEEE, 2017.

\bibitem{laina2016deeperdepth}
I.~Laina, C.~Rupprecht, V.~Belagiannis, F.~Tombari, and N.~Navab.
\newblock Deeper depth prediction with fully convolutional residual networks.
\newblock In {\em 3D Vision (3DV), 2016 Fourth International Conference on},
  pages 239--248. IEEE, 2016.

\bibitem{liu2015depth}
F.~Liu, C.~Shen, and G.~Lin.
\newblock Deep convolutional neural fields for depth estimation from a single
  image.
\newblock In {\em Proceedings of the IEEE Conference on Computer Vision and
  Pattern Recognition}, pages 5162--5170, 2015.

\bibitem{liu2016depth}
F.~Liu, C.~Shen, G.~Lin, and I.~Reid.
\newblock Learning depth from single monocular images using deep convolutional
  neural fields.
\newblock {\em IEEE transactions on pattern analysis and machine intelligence},
  38(10):2024--2039, 2016.

\bibitem{long2015fcn}
J.~Long, E.~Shelhamer, and T.~Darrell.
\newblock Fully convolutional networks for semantic segmentation.
\newblock In {\em Proceedings of the IEEE Conference on Computer Vision and
  Pattern Recognition}, pages 3431--3440, 2015.

\bibitem{LonguetHiggins81}
H.~C. Longuet-Higgins.
\newblock A computer algorithm for reconstructing a scene from two projections.
\newblock {\em Nature}, 293(5828):133--135, 1981.

\bibitem{lowe2004sift}
D.~G. Lowe.
\newblock Distinctive image features from scale-invariant keypoints.
\newblock {\em International journal of computer vision}, 60(2):91--110, 2004.

\bibitem{mur2015orbslam}
R.~Mur-Artal, J.~M.~M. Montiel, and J.~D. Tardos.
\newblock Orb-slam: a versatile and accurate monocular slam system.
\newblock {\em IEEE Transactions on Robotics}, 31(5):1147--1163, 2015.

\bibitem{Silberman2012nyuv2}
P.~K. Nathan~Silberman, Derek~Hoiem and R.~Fergus.
\newblock Indoor segmentation and support inference from rgbd images.
\newblock In {\em ECCV}, 2012.

\bibitem{noh2015learning}
H.~Noh, S.~Hong, and B.~Han.
\newblock Learning deconvolution network for semantic segmentation.
\newblock In {\em Proceedings of the IEEE International Conference on Computer
  Vision}, pages 1520--1528, 2015.

\bibitem{ronneberger2015u}
O.~Ronneberger, P.~Fischer, and T.~Brox.
\newblock U-net: Convolutional networks for biomedical image segmentation.
\newblock In {\em International Conference on Medical Image Computing and
  Computer-Assisted Intervention}, pages 234--241. Springer, 2015.

\bibitem{rublee2011orb}
E.~Rublee, V.~Rabaud, K.~Konolige, and G.~Bradski.
\newblock Orb: An efficient alternative to sift or surf.
\newblock In {\em Computer Vision (ICCV), 2011 IEEE international conference
  on}, pages 2564--2571. IEEE, 2011.

\bibitem{ILSVRC15}
O.~Russakovsky, J.~Deng, H.~Su, J.~Krause, S.~Satheesh, S.~Ma, Z.~Huang,
  A.~Karpathy, A.~Khosla, M.~Bernstein, A.~C. Berg, and L.~Fei-Fei.
\newblock {ImageNet Large Scale Visual Recognition Challenge}.
\newblock {\em International Journal of Computer Vision (IJCV)},
  115(3):211--252, 2015.

\bibitem{saxena2006depth}
A.~Saxena, S.~H. Chung, and A.~Y. Ng.
\newblock Learning depth from single monocular images.
\newblock In {\em Advances in neural information processing systems}, pages
  1161--1168, 2006.

\bibitem{saxena2009make3d}
A.~Saxena, M.~Sun, and A.~Y. Ng.
\newblock Make3d: Learning 3d scene structure from a single still image.
\newblock {\em IEEE transactions on pattern analysis and machine intelligence},
  31(5):824--840, 2009.

\bibitem{song2014robust}
S.~Song and M.~Chandraker.
\newblock Robust scale estimation in real-time monocular sfm for autonomous
  driving.
\newblock In {\em Proceedings of the IEEE Conference on Computer Vision and
  Pattern Recognition}, pages 1566--1573, 2014.

\bibitem{triggs1999bundle}
B.~Triggs, P.~F. McLauchlan, R.~I. Hartley, and A.~W. Fitzgibbon.
\newblock Bundle adjustment—a modern synthesis.
\newblock In {\em International workshop on vision algorithms}, pages 298--372.
  Springer, 1999.

\bibitem{ummenhofer2016demon}
B.~Ummenhofer, H.~Zhou, J.~Uhrig, N.~Mayer, E.~Ilg, A.~Dosovitskiy, and
  T.~Brox.
\newblock Demon: Depth and motion network for learning monocular stereo.
\newblock In {\em IEEE Conference on Computer Vision and Pattern Recognition
  (CVPR)}, 2017.

\bibitem{vijayanarasimhan2017sfmnet}
S.~Vijayanarasimhan, S.~Ricco, C.~Schmid, R.~Sukthankar, and K.~Fragkiadaki.
\newblock Sfm-net: Learning of structure and motion from video.
\newblock {\em arXiv preprint arXiv:1704.07804}, 2017.

\bibitem{wang2017deepvo}
S.~Wang, R.~Clark, H.~Wen, and N.~Trigoni.
\newblock Deepvo: Towards end-to-end visual odometry with deep recurrent
  convolutional neural networks.
\newblock In {\em Robotics and Automation (ICRA), 2017 IEEE International
  Conference on}, pages 2043--2050. IEEE, 2017.

\bibitem{wang2004image}
Z.~Wang, A.~C. Bovik, H.~R. Sheikh, and E.~P. Simoncelli.
\newblock Image quality assessment: from error visibility to structural
  similarity.
\newblock {\em IEEE transactions on image processing}, 13(4):600--612, 2004.

\bibitem{weerasekera2017feature}
C.~S. Weerasekera, R.~Garg, and I.~Reid.
\newblock Learning deeply supervised visual descriptors for dense monocular
  reconstruction.
\newblock {\em arXiv preprint arXiv:1711.05919}, 2017.

\bibitem{ye2017self}
M.~Ye, E.~Johns, A.~Handa, L.~Zhang, P.~Pratt, and G.-Z. Yang.
\newblock Self-supervised siamese learning on stereo image pairs for depth
  estimation in robotic surgery.
\newblock {\em arXiv preprint arXiv:1705.08260}, 2017.

\bibitem{yi2016lift}
K.~M. Yi, E.~Trulls, V.~Lepetit, and P.~Fua.
\newblock Lift: Learned invariant feature transform.
\newblock In {\em European Conference on Computer Vision}, pages 467--483.
  Springer, 2016.

\bibitem{zbontar2016mccnn}
J.~Zbontar and Y.~LeCun.
\newblock Stereo matching by training a convolutional neural network to compare
  image patches.
\newblock {\em Journal of Machine Learning Research}, 17(1-32):2, 2016.

\bibitem{zhou2017sfmlearner}
T.~Zhou, M.~Brown, N.~Snavely, and D.~G. Lowe.
\newblock Unsupervised learning of depth and ego-motion from video.
\newblock In {\em CVPR}, 2017.

\end{thebibliography}

}

\end{document}